\documentclass[journal=jacsat,manuscript=article]{achemso}

\usepackage{amsmath}
\usepackage{amsbsy}
\usepackage{booktabs}
\usepackage{xcolor}
\usepackage{graphicx}
\usepackage{soul,color}
\usepackage{titlesec}
\usepackage{cleveref}
\usepackage{array}
\usepackage{float}
\usepackage{enumitem}
\usepackage[normalem]{ulem}
\newcommand{\bs}[1]{\boldsymbol{#1}}

\newcommand{\mc}[1]{\mathcal{#1}}

\author{Ayush Jain$^\star$}
\affiliation{Carbon Inc., Redwood City, CA}
\alsoaffiliation{Georgia Institute of Technology, School of Materials Science and Engineering, Atlanta, GA}
\email{ayush.jain@gatech.edu}
\author{Ehsan Haghighat$^\star$}
\affiliation{Carbon Inc., Redwood City, CA}
\email{ehsan_haghighat@carbon3d.com}
\author{Sai Nelaturi}
\affiliation{Carbon Inc., Redwood City, CA}
\email{snelaturi@carbon3d.com}

\title
{LatticeGraphNet: A two-scale graph neural operator for simulating lattice structures}

\begin{document}

% \begin{tocentry}

% Some journals require a graphical entry for the Table of Contents.
% This should be laid out ``print ready'' so that the sizing of the

% Inside the \texttt{tocentry} environment, the font used is Helvetica
% 8\,pt, as required by \emph{Journal of the American Chemical

% The surrounding frame is 9\,cm by 3.5\,cm, which is the maximum
% permitted for  \emph{Journal of the American Chemical Society}
% graphical table of content entries. The box will not resize if the
% content is too big: instead it will overflow the edge of the box.

% This box and the associated title will always be printed on a
% separate page at the end of the document.

% \end{tocentry}

\begin{abstract}
% Additive manufacturing has facilitated the adoption of elastomeric meta-materials with tunable mechanical characteristics for use in many products. However, designing meta-materials for novel applications requires the characterization of a large design space. Finite element simulations assist this search by approximating the compressive response of elastomeric meta-materials, albeit with extensive computational complexity and runtime. 
This study introduces a two-scale Graph Neural Operator (GNO), namely, LatticeGraphNet (LGN), designed as a surrogate model for costly nonlinear finite-element simulations of three-dimensional latticed parts and structures. LGN has two networks: LGN-i, learning the reduced dynamics of lattices, and LGN-ii, learning the mapping from the reduced representation onto the tetrahedral mesh. LGN can predict deformation for arbitrary lattices, therefore the name operator. 
Our approach significantly reduces inference time while maintaining high accuracy for unseen simulations, establishing the use of GNOs as efficient surrogate models for evaluating mechanical responses of lattices and structures.
\end{abstract}

\paragraph{Keywords}{Graph Neural Networks; Neural Operators; Structural Analysis; Metamaterials.}
% \keywords{AAAA}
\def\thefootnote{$\star$}\footnotetext{AJ and EH contributed equally to this work.}

% \newpage
% \section*{Highlights}
% \begin{enumerate}
%     \item LatticeGraphNet is introduced for simulating lattices and truss and beam structures. 
%     \item Once trained, LGN can predict deformation for arbitrarily meshed structures. 
%     \item LGN offers a two-scale approach to increase accuracy. 
%     \item LGN predicts nonlinear and buckling deformations. 
% \end{enumerate}
% \newpage

% \newpage
% \tableofcontents

% ------------------
\section{Introduction}

% \hl{Add an image of lattices from Ruiqi's work}\\

Lattice structures, 
% \sout{with their unique repetitive patterns} 
{constructed by repeated copies and interpolation of smaller scale truss structures}, are crucial in diverse scientific fields.
% \sout{, especially additive manufacturing}
In materials science, they are known for their exceptional strength-to-weight ratio, ideal for applications requiring both lightness and durability \cite{deep-meta, Garland-genopt, wu2021brief}. These structures are also mirrored in biological systems, illustrating nature's efficiency in structural design.
The rise of 3D printing has particularly highlighted their significance, enabling the creation of 
% \sout{complex, customized geometries} 
{customized designs with complex geometries and spatially tuned material properties}\cite{jiao2023mechanical, Carbon_MML} once thought unfeasible \cite{DLP_review, DLP_review2}. 
% \sout{This innovation is}
{Such innovations are} vital in sectors like aerospace and biomedical engineering, where tailored lattice designs can meet specific needs such as improved load-bearing capabilities or enhanced thermal conductivity. Thus, lattice structures represent a key intersection of theory and practical application across various scientific {and engineering} disciplines.

{Physical experimentation is often employed} to understand the mechanical characteristics of lattice structures.
% \sout{, physical experimentation is often employed}
This requires {fabricating} the structure first 
{(typically with 3D printing)} and conducting physical testing, which are both costly. To reduce the cost of physical characterization, high-fidelity numerical simulations, using the Finite Element Method (FEM), have been used to guide design \cite{high-thru-design, high-thru-design-2}. 
For example, our team at Carbon Inc. has demonstrated that the Incremental Potential Contact (IPC)\cite{IPC} method can accurately characterize the compressive response of elastomeric lattice structures by accounting for large deformations, instabilities (buckling) and contact \cite{Carbon_MML}. 
However, the complexity of IPC simulations results in a long runtime, on the order of \textbf{48 hours} to \textbf{10 days} per model, depending on the mesh size, topological characteristics, and nonlinearities due to buckling and self-contact. Therefore, we experience the need for faster simulations firsthand. 

\subsection{Machine Learning To Accelerate Physical Simulation}
\label{Sec:ML-Physics}
In recent years, machine-learning (ML) techniques have made major inroads to accelerate physical simulations \cite{ML1,ML2,ML3,ML4}. 
There have been some efforts to offer ML-based alternatives to traditional boundary value problem (BVP) solvers, e.g., Physics-Informed Neural Networks (PINNs) \cite{PINNs, PINN-solid, PINN-Gerorge}.
% {\color{red}\cite{sukumar2022exact, berrone2022enforcing}}
They have also been used in generative design \cite{Garland-genopt, zheng2023unifying} and inverse design \cite{ha2023rapid, deep-meta, ma2019probabilistic, kumar2022inverting, tran2019constrained} of lattices and meta-materials. 
% Despite the progress, the characterization of industrial component-sized and hybrid unit cell lattices remains a challenge.
Despite the progress, the application of ML solvers in realistic (industrial) applications remains very limited.
Recently, Graph Neural Networks (GNNs) have been considered the state-of-the-art for learning physical phenomena \cite{sanchez2020learning, lino2021simulating, han2022predicting, guo2020semi}.
Specifically, an ML architecture called MeshGraphNet (MGN) \cite{MGN, multiscaleMGN} is proposed to be a generalizable GNN architecture that can learn the dynamics of mesh-based simulations. MGNs work by encoding the discretization mesh into a multigraph of nodes and edges. This technique boasts short inference times with low point-by-point error comparisons between finite-element simulation counterparts \cite{MGN}. Given this strength, MGN-based architectures have the potential to aid in the characterization of meta-material components. 

\subsection{Contributions}
\label{Sec:Contributions}
% {\color{blue}
% \hl{The work in this study and its contributions...}\\

\begin{figure*}[h]
    \centering
    \includegraphics[width=0.8\columnwidth]{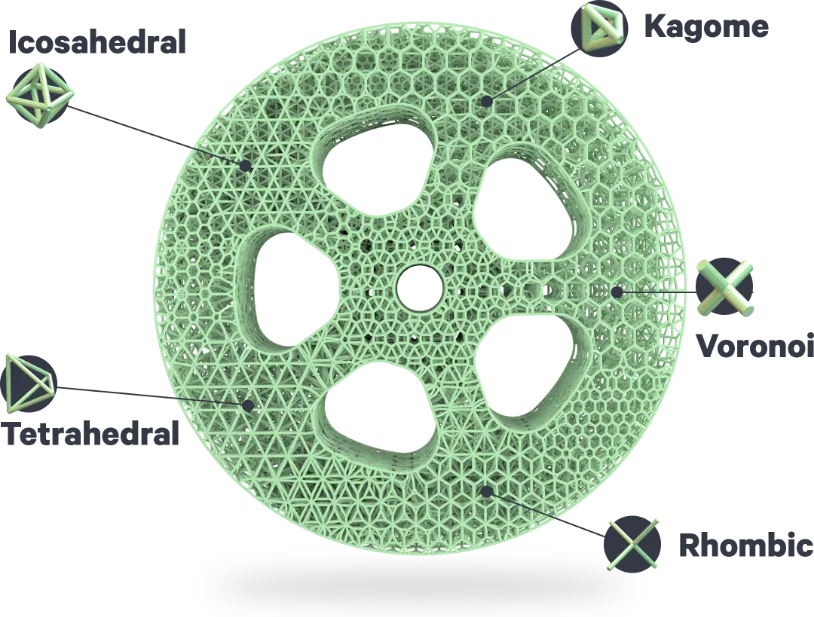}
    \caption{A sample 3D printed elastomeric part with a combination of five different unit cells, demonstrating the variety of configurations achieved by a small number of initial repeat blocks.\cite{Carbon_lattice}}
    \label{fig:lattices}
\end{figure*}

\begin{figure*}[h]
    \centering
    \includegraphics[width=\columnwidth]{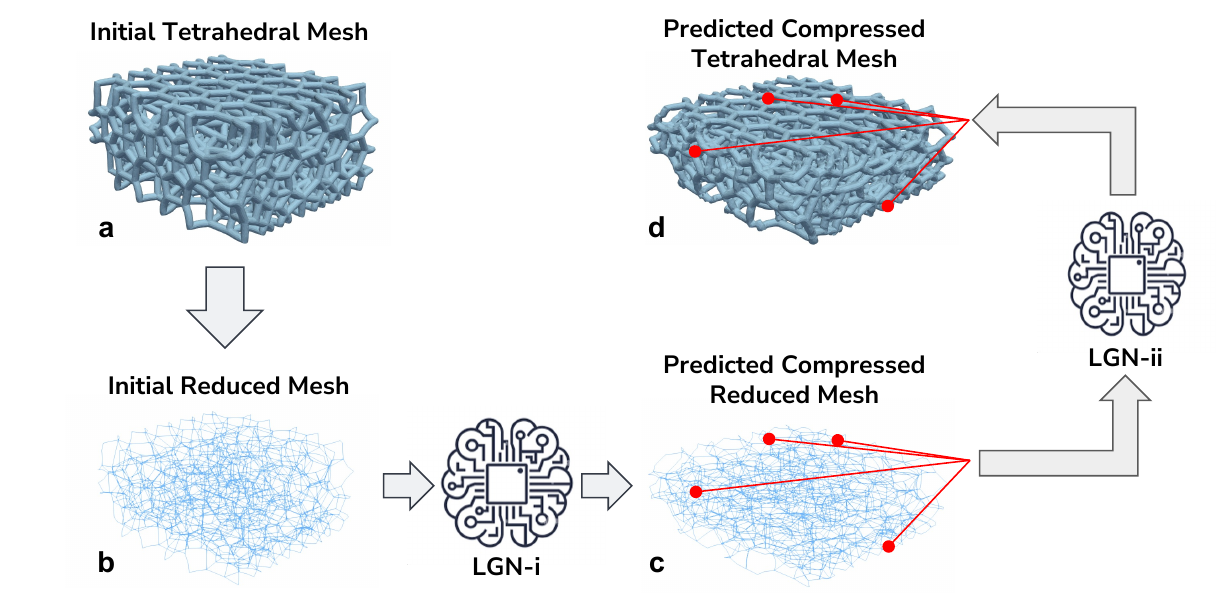}
    \caption{Overview of the LatticeGrapNet (LGN) Pipeline. The pipeline starts with A) an initial three-dimensional lattice represented by a tetrahedral mesh and is transformed to B) a reduced (skeletal) representation of the initial mesh. C) LGN-i (see \Cref{sec:lgn-i}) runs inference on the reduced mesh to get the coarse displacement. D) LGN-ii (see \Cref{sec:lgn-ii}) maps the reduced displacements of LGN-i to predict three-dimensional displacements on the tetrahedral mesh.
    }
    \label{fig:overview}
\end{figure*}

In this work, we propose LatticeGraphNet (LGN), an MGN architecture (detailed in \Cref{fig:overview}) that serves as a surrogate model for high-fidelity nonlinear Neohookean IPC simulations of three-dimensional latticed meta-materials (see \Cref{fig:lattices}). 
Latticed components used for IPC simulations typically require the use of high-precision discretizations.
These representations are too complex (high-dimensional) for the standard MGN model, making it difficult for any GNN algorithm to learn directly from them. Our multi-scale LGN architecture overcomes this complexity by using two MGN-based architectures, each predicting the dynamics on different precision levels.
The input to the pipeline is a complex three-dimensional tetrahedral mesh of a latticed part. The first MGN-based model, LGN-i, encodes a reduced (beam) representation of the tetrahedron mesh by leveraging the skeletal representation of the tetrahedral mesh. The predicted reduced displacements serve as information for the second MGN-based architecture, LGN-ii, to predict the full-scale three-dimensional displacements on the tetrahedral points. This allowed us to use only 108 IPC simulations to train the model with good accuracy. It is worth noting that the total cost of FEM simulations for these models is approximately 8,000~hr.
% \sn{-- do we discuss the time it took to generate the dataset?}

We find that our approach reduces the time of running inference to \textbf{10 to 30 minutes}, depending on the size of the tetrahedral mesh. For unseen simulations, our approach maintains an average point-by-point accuracy within 1.67\% of the structure size.
%maintains an average point-by-point accuracy of <1 mm for unseen simulations with a bounding box diagonal of 60mm. 
% \sn{I find the 0.389 a bit weird and specific, is it agnostic of the scale of the training data geometries? In our case all the training is done on lattices of a particular size, which makes the 0.389 relevant to those types of lattices, and not universally across all truss structures}
This work establishes a baseline for the use of GNNs as surrogate models to simulate the mechanical responses of lattices and structures.
% }

% In this work, we propose a GNN pipeline (detailed in Figure 1) that serves as a surrogate model for Neohookean IPC simulations of high-precision latticed meta-materials. Our pipeline overcomes the complexity of our representations by using two MGN-based architectures, each predicting the dynamics on different precision levels. The input to the pipeline is a complex tetrahedron mesh of a latticed metamaterial. Our first MGN-based model, called LatticeGraphNet (LGN), encodes a coarse representation of the tetrahedron mesh into a multi-graph, and estimates the coarse displacements of components under compression. The predicted coarse displacements serve as information for the second MGN-based architecture, called TetGraphNet (TGN), to predict the fine displacements within tetrahedrons of the original mesh.

% We find that our approach reduces the time of running inference to \textbf{minutes} and maintains a point-by-point accuracy of --- for unseen simulations. As a baseline, this work establishes the use of GNNs to be used as surrogate models to simulate the mechanical responses of meta-materials, and can be used to rapidly evaluate large sets of meta-materials.

% -------------------

\section{Dynamics of lattices}
Let us consider the undeformed (reference) coordinate of a material point at time $t$ as $\bs{X}$. The deformed (spatial) coordinates of the same point are then expressed as $\bs{x} = \phi(\bs{X}, t)$ with $\phi$ as the mapping between reference and spatial frames. This mapping can also be expressed using the displacement vector $\bs{u}$ as $\bs{\phi}(\bs{X},t) = \bs{u}(\bs{X},t) + \bs{X}$. The Jacobian of this mapping is called the deformation gradient tensor $\bs{F}$ and is expressed as 
\begin{align}\label{eqs:defgrad}
    \bs{F} &= \nabla \bs{\phi} = \nabla \bs{u} + \bs{I},
\end{align}
in which $\nabla = \frac{\partial}{\partial \bs{X}}$ is called the \emph{right} reference gradient. \cite{MatPointMethod}

We consider elastomeric materials, and their mechanical response is described using the hyperelastic Neo-Hookean model under large deformations. The Neo-Hookean stored energy function is expressed as 
\begin{align}
    \Psi(\bs{F}) = \frac{1}{2} \lambda (\ln J)^2 - \mu \ln J + \frac{1}{2}\mu (\bs{F}:\bs{F} - 3),
\end{align}
where $J$ is the volumetric deformation of the material and expressed $J=\det(\bs{F})$ and $\mu$ and $\lambda$ are Lam\'e parameters. The first Piola-Kirchhoff (PK) stress tensor $\bs{P}$ is then derived as 
\begin{align}\label{eqs:pk-stress}
    \bs{P} = \frac{\partial \Psi}{\partial \bs{F}} = \lambda \ln(J) \bs{F}^{-T} + \mu (\bs{F} - \bs{F}^{-T}).
\end{align}
The governing equations describing this system are mass conservation, expressed as $\rho J = \rho_0$, and linear momentum conservation, expressed as 
\begin{align}\label{eq:equation_of_motion}
    \nabla \bs{P} + \bs{b} = \rho_0 \frac{\partial^2 \bs{u}}{\partial t^2},
\end{align}
where $\bs{b}$ is the body force vector and $\rho_0$ is density of material at its undeformed state. The second (angular) momentum conservation relation results in the symmetry condition on the stress tensor, i.e., $\bs{F}\bs{P} = (\bs{F}\bs{P})^T$\cite{MatPointMethod}. Note that we will use mean and deviatoric invariants of the PK stress tensor, i.e., $I_1 = tr(\bs{P})$ and $J_2 = \frac{1}{2} \bs{P}:\bs{P}$, respectively, as state features in training the graph neural networks. 

The training data for the deformation of lattices are generated based the NeoHookean elastodynamic formulation described above, using the IPC method \cite{IPC}.

\section{Lattice Graph Network}
\label{Sec:LGN}

\emph{Neural operators} are designed to learn the generative solution of parametric PDEs, making them exceptionally suited for dealing with complex simulations often encountered in physics and engineering \cite{NO1,NO2,NO3,NO4,NO5}.
This growing interest is reflected in a surge of research and publications as scientists and engineers seek to leverage neural operators for more accurate, real-time predictions and decision-making processes. 
MeshGraphNet (MGN) \cite{MGN} is a class of graph neural operators that is particularly relevant here for modeling lattices and structures. The inputs to the MGN architecture include the nodes and edges of the simulation mesh, nodal quantities of interest (QoI) at time $t$, as well as any parameterization of the system, and the outputs are QoI increments, such as displacement and pressure increments, at mesh nodes. 
Therefore, the network learns the physics of interactions between nodes based on edge connectivities. It is a generative model, meaning that once trained, we can input any mesh data and expect predictions for the QoI. 

In this work, we build a surrogate model of lattices in three dimensions, like those shown in \Cref{fig:overview}, using a small number of high-fidelity simulation data.
It is worth noting that the reduced (beam) representation of such structures {is sufficient to predict deformations}; however, we choose {to build a surrogate model of} the three-dimensional representation so that volumetric stresses are naturally measurable. 
We find that the standard MGN architecture fails to learn the response of the system, partly due to a small number of training samples and partly due to the very structure of the mesh {(which itself is a particular discretization of the design)}. Note that some edges of the tetrahedral mesh do represent the underlying beam-like structure, e.g., those coincide with the skeletal representation, and others behave very differently. We hypothesize that this issue is the root cause of the challenges we faced with {directly} using the standard MGN architecture. Hence, we propose a two-scale sequential approach:
\begin{enumerate}[label=(\roman*)]
    \item An MGN model predicting the reduced (beam) representation of the three-dimensional lattice.
    \item An MGN model predicting the mapping between the reduced representation and the three-dimensional representation. 
\end{enumerate}
These models are detailed below. Note that to differentiate with three-dimensional state variables, those for the reduced representation are shown with $\tilde{\circ}$, e.g., $\tilde{\bs{u}}$ for lattice displacements.

% \hl{discuss why we need to predict stresses. AJ - added in decoder section}
% \hl{provide further discussions on the two-scale approach. }
% \hl{discuss objectives: learning nonlinear buckling etc}

\subsection{LGN-i: The reduced predictor}\label{sec:lgn-i}

% \subsubsection{LGN Architecture}

\begin{figure*}[h]
    \centering
    \includegraphics[width=\columnwidth]{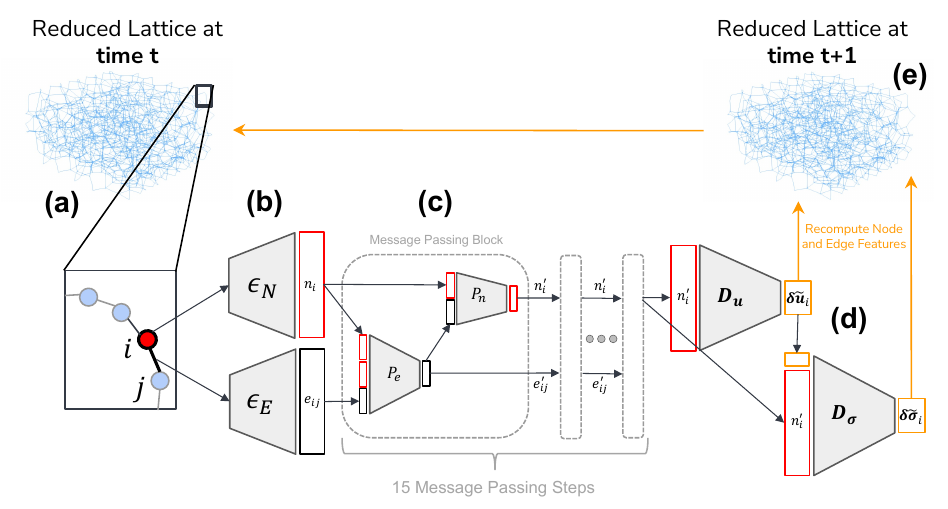}
    \caption{Architecture of LGN-i. 
    A) The input to the network is a reduced (graph) representation of the tetrahedral mesh. 
    B) The node encoder ($\epsilon_N$) and edge encoder ($\epsilon_E$) encode the node and edge features to high-dimensional vectors, $n_i$ and $e_{ij}$, respectively. 
    C) The message-passing block uses two \emph{processor} MLPs: $P_e$ which updates $e_{ij}$ to $e'_{ij}$, and $P_n$ which updates $n_i$ to $n'_i$. LGN-i uses 15 message-passing steps with identical blocks. 
    D) The two-part decoder contains MLPs $D_{\bs{u}}$ and $D_{\bs{\sigma}}$. $D_{\bs{u}}$ uses the updated node features to compute the displacement increment for node $i$, i.e., $\delta\tilde{\bs{u}}_i$. $D_{\sigma}$ uses the final $n'_i$ concatenated with $\bs{u}_i$ to calculate the change in the $I_1$ and $J_2$ stress invariants at node $i$, $\delta \tilde{\bs{\sigma}}_i$. 
    E) Upon rollout, denoted by the orange lines, the calculated $\delta \tilde{\bs{u}}_i$ and $\delta \tilde{\bs{\sigma}}_i$ are integrated to obtain the node and edge features for the next timestep.
    }
    \label{fig:LGN_arch}
\end{figure*}

The reduced MGN network, namely LGN-i, detailed in \Cref{fig:LGN_arch}, contains three main parts: an encoder, a processor, and a decoder, similar to the original MeshGraphNet \cite{MGN}. The architecture is an autoregressive type, meaning at every time step, the inputs to the network are the state parameters at time $t$ and the outputs include the state parameters at the next time step $t+1$. Hence, given the state of a strut lattice at time $t$, i.e., $L_t(N,E)$ with nodes $N$ and edges $E$, the network predicts the displacement increment $\delta\tilde{\bs{u}}$ of each node at time $t+1$. 

% The input state variables of choice are the current nodal position $\boldsymbol{x}_i(t) = (x_i, y_i, z_i)$, and the nodal mean ($I_1$) and deviatoric ($J_2$) stress components.
% % of the part $V$ are $\boldsymbol{\sigma}_i(t) = (I1_i, J2_i)$ (discussed below). 

% The overall Boundary Value Problem can be written as \\
% $$\partial_{t+1} \boldsymbol{x}_i = f(L_t(V,E))$$
% Initial Condition: $$\boldsymbol{x}_i(0) = L_0(V,E)$$
% Initial Condition: $$\boldsymbol{\sigma}_i(0) = (0, 0)$$
% Boundary Condition: $$\partial_{t+1} z_i = 0 : z_i = 0$$ 
% Boundary Condition: $$\partial_{t+1} z_i = v : z_i = max(z_i)$$

% Where $f$ is the Partial Differential Equation (PDE) learned by the LGN \cite{}, and $v$ is the per timestep velocity of compression on the upper nodes of the lattice, denoted by $max(z_i)$.

\paragraph{Encoder}
The input features of nodes and edges contain physical attributes that represent the current state of the lattice structure.
% Using these features, the encoder encodes the lattice nodes and edges into graph vertices and bidirectional edges.
Node features consist of:
\begin{itemize}
    \item Node-type: a one-hot encoded node type vector of size 4, representing fixed boundary nodes, loading boundary nodes, free boundary nodes, and interior nodes,
    % \item Node-type: a one-hot encoded node type vector of size 4, representing one of upper boundary node, lower boundary node, side boundary node, and inside node 
    % \sn{this is confusing to me -- a node is a point, so what is an upper/lower node encoding -- explaining this just a little more will be very helpful},
    \item Strut diameter,
    \item Node degree, indicating the number of edge connections with other nodes in the mesh,
    \item Mean and deviatoric stress invariants ($\tilde{I}_1$, $\tilde{J}_2$).
\end{itemize}
We found that adding stress information as state variables helps with learning the buckling behavior. 
Edge features include information about the displacement between nodes:
% We represent spatial information through the “mesh space,” which encodes a representation of the original displacements between the connected nodes $i$ and $j$ ($u_{ij}$), and the “world space,” which represents the Euclidean distance between the nodes during the current state ($w_{ij}$).}

\begin{itemize}
    \item ``Reference-space" which encodes the \emph{undeformed} distance between connected nodes (i.e., $\tilde{\bs{X}}_i - \tilde{\bs{X}}_j$ for the edge connecting nodes $i$ and $j$),
    \item ``Current-space" which encodes the \emph{deformed} distance between connected nodes (i.e., $\tilde{\bs{x}}_i - \tilde{\bs{x}}_j$ for the edge connecting nodes $i$ and $j$).  
\end{itemize}
Note that each {edge} feature has a size of four: three for the vector components and one for the length of the vector. As shown in \Cref{fig:LGN_arch}, there are two encoder MLPs, $\epsilon_N$ for encoding nodal features and $\epsilon_E$ for encoding edge features. The encoders map their respecting node and edge features to a latent vector of size 128.

\paragraph{Processor} The processor contains 15 sequential message-passing blocks. Each block contains two MLPs with residual connections, $P_n$ and $P_e$, mapping input node embeddings $n_i$ and edge embeddings $e_{ij}$ to their updated corresponding node and edge embeddings $n'_i$ and $e'_{ij}$, respectively. Each message-passing block is expressed as 
\begin{equation}
    e'_{ij} = P_e(e_{ij}, n_i, n_j), \quad n'_{i} = P_n(\sum_j e'_{ij}, n_i).
\end{equation}

\paragraph{Decoder} The decoder maps the latent node embeddings $n'_{i}$ to final displacement and stress increments through the displacement decoder $D_{\bs{u}}$ and stress decoder $D_{\bs{\sigma}}$ networks. We also append the output of the displacement decoder to the input of the stress decoder network. This is expressed as 
\begin{equation}
    \delta\tilde{\bs{u}}_i = D_{\bs{u}}(n'_i), \quad \delta\tilde{\bs{\sigma}}_i = D_{\bs{\sigma}}(n'_i, \delta\tilde{\bs{u}_i}).
\end{equation}
Note that predicting stress components $I_1$ and $J_2$ are needed since they are state variables.

\paragraph{Training}
To train the network, we follow a sequential approach. We first train the network only on displacements. Then, by freezing the parameters of the encoder, processor, and displacement decoder, we optimize the parameters of the stress-decoder network. Our reasoning is that we want the network to have the highest possible accuracy on displacements. 
To even further improve the accuracy of displacement predictions, we also adopt a strategy inspired by the ``pushforward loss" technique \cite{MPNN-PDE}. Here, LGN undergoes a rollout for two time-steps starting from $L_t(N,E)$. After the first timestep prediction, we truncate the computational graph to stop further backpropagation due to high memory requirements. 
Hence, the push-forwarded loss is only back-propagated through the second pass of the LGN. Therefore, the final loss functions are
\begin{align}
\mc{L}_{\bs{u}} &= \frac{1}{N} \sum_i \left(\tilde{\bs{u}}^i - \tilde{\bs{u}}^{i*}\right)^2 + \frac{1}{N}\sum_i \left(\tilde{\bs{u}}^i_{t+2} - \tilde{\bs{u}}^{i*}_{t+2}\right)^2, \\
\mc{L}_{\sigma} &= \frac{1}{N} \sum_i \left(\tilde{\bs{\sigma}}^i - \tilde{\bs{\sigma}}^{i*}\right)^2,
\end{align}
where $\circ^*$ represents the training data. Note that the training data for the reduced dimension is evaluated by interpolating the three-dimensional data (see \Cref{sec:dataset} for more details). 
Compared to a coupled standard optimization of both variables without the push-forward loss, we found that this strategy increases the accuracy of both displacement and stress predictions.

% \sn{Something is missing here -- how do we obtain the $\tilde{\bs{u}}^{i*}$? There should be a mapping from the IPC results on the tet mesh to the beam representation in the training data. Where do we talk about that?}

\subsection{LGN-ii: The tetrahedral predictor}\label{sec:lgn-ii}
% \subsection{TetGraphNet}

\begin{figure*}[h!]
    \centering
    \includegraphics[width=\columnwidth]{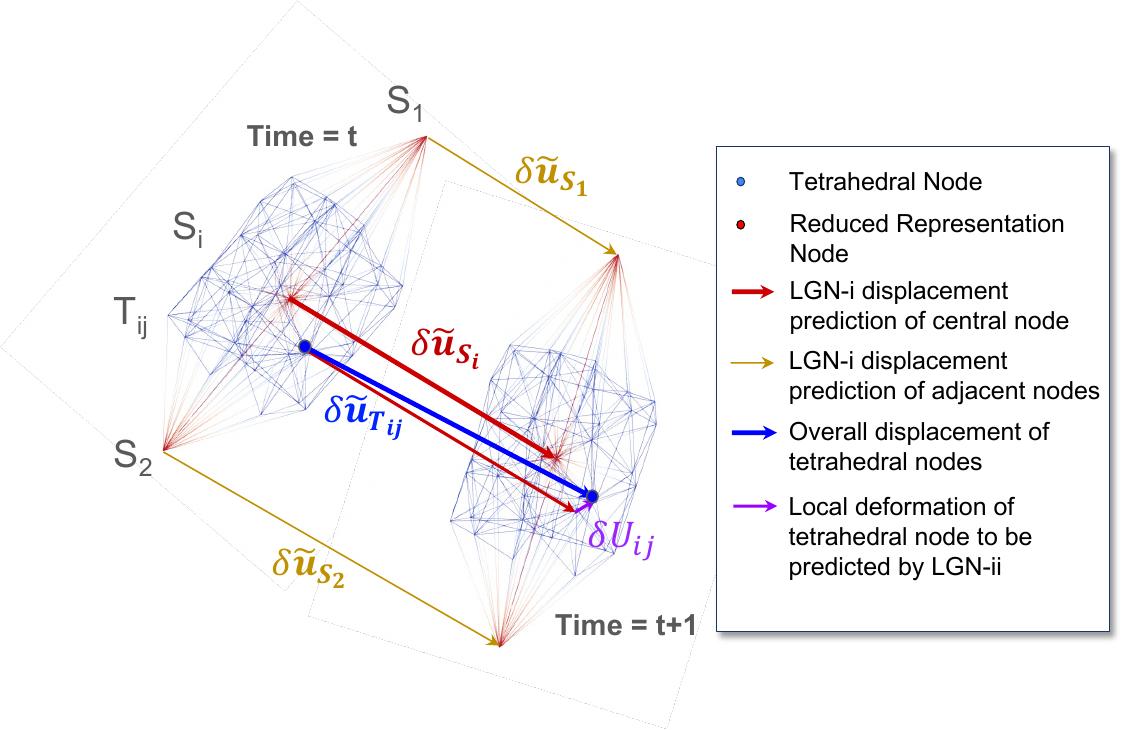}
    \caption{Volumetric deformation prediction set up by the LGN-ii. At time $t$, an initial configuration of the $i^{th}$ strut node ($S_i$), its nearby tetrahedral nodes ($T_{ij}$), and its connected strut nodes ($S_1$, $S_2$) are known.
    LGN-i predicts the displacement of every strut node ($\delta \tilde{\bs{u}}_{S_i}$, $\delta \tilde{\bs{u}}_{S_1}$, $\delta \tilde{\bs{u}}_{S_2}$). Based on this information, LGN-ii predicts the fine local deformation of the $j^{th}$ tetrahedral node ($\delta \bs{U}_{ij}$). The final tetrahedral node displacement is $\delta \bs{u}_{T_{ij}} = \delta \bs{U}_{ij} + \delta \tilde{\bs{u}}_{S_i}$. }
    \label{fig:LGNii_arch}
\end{figure*}

As discussed above, our objective is to predict the complete three-dimensional volumetric response of the lattice structure; this allows us to estimate homogenized stresses for lattices. Since the first network learns the bulk deformation on the reduced mesh, we need a second network to map the reduced deformations onto its corresponding volumetric mesh, hence the LGN-ii here. Instead of predicting the displacement of the complete volume at once, LGN-ii iterates over each node in the reduced strut lattice ($S_i$). For every $S_i$ we build a sub-graph to encode spatial information between itself, its neighboring nodes ($S_k$), and its assigned volumetric tetrahedral nodes ($T_{ij}$). This sub-graph is used as the input to LGN-ii to predict the deformation of nodes $T_{ij}$ ($\delta \bs{u}_{T_{ij}}$) based on the LGN-i prediction of $\delta \tilde{\bs{u}}_{S_{k}}$ and $\delta \tilde{\bs{u}}_{S_{i}}$. This problem setup is illustrated in \Cref{fig:LGNii_arch}. Similar to LGN-i, LGN-ii is also an MGN network with an encoder, a processor, and a decoder.

\paragraph{Encoder} LGN-ii uses three encoder blocks: 
\begin{itemize}
    \item strut node encoder ($\epsilon_{SN}$), which encodes ($S_i$, $S_k$) to ($sn_i$, $sn_k$),
    \item tetrahedral node encoder ($\epsilon_{TN}$), which encodes the state of $T_{ij}$ to $tn_i$,
    \item edge encoder ($\epsilon_{TNE}$) encodes strut and tetrahedral edges ($E_c$) to $e_c$.
\end{itemize}
($S_i$, $S_k$) are represented by a feature vector of size 4, which includes the strut diameter and the predicted displacement relative to the $i^{th}$ strut node ($\delta \bs{\tilde{u}}_{S_{k}} - \delta \bs{\tilde{u}}_{S_{i}}$). $T_{ij}$ nodes are provided a single feature of strut diameter. $E_c$ edges are provided with edge type as a one-hot encoded feature, and the Euclidean distance and magnitude between the connected nodes.

\paragraph{Processor} Similar to the LGN-i, message-passing blocks are comprised of two MLPs, $P_{Te}$ for edge updates and $P_{Tn}$ for node updates, with a hidden layer size of 64. Only four message-passing blocks are used in the LGN-ii. The blocks update $tn_i$ to $tn'_i$.

\paragraph{Decoder} The final block is the decoder block $D_U$, which maps $tn'_i$ to the relative incremental displacements on the tetrahedral mesh $\delta \bs{U}_{ij}$. 

% \paragraph{Training} \hl{Provide details about training the network ...}

\subsection{Post-processing and homogenization}

The FEM discretization provides a straightforward mechanism to calculate reaction forces.
However, we cannot directly calculate reaction forces from LGN outputs. Instead, we need to use LGN displacement predictions of the tetrahedral meshes to predict element deformation gradient and PK-stress components using \Cref{eqs:defgrad,eqs:pk-stress} respectively. Lastly, using cross-sectional equilibrium, we can approximate reaction forces, as shown in \Cref{fig:Homogenization}. 
Horizontal slices are taken through the mesh at heights of 0.1, 0.15, 0.20, 0.25, 0.5, and 0.75. Multiple slices are used to mitigate approximation errors. 
The total  traction force on each cross-section $\Gamma_i$ is then given by
\begin{equation}
F_i = \int_{\Gamma_i} \bs{P}\cdot \bs{n}~dA.
\end{equation}
The homogenized reaction force is ultimately measured as the average of $F_i$ on all cross-sections.

\section{Results and discussion}
In this section, we first summarize the details regarding the training dataset and model training and then present LGN results on test (unseen) structures. As per the implementation, we implemented and trained LGN on Nvidia's Modulus \cite{modulus} by revising their MGN implementation. 

\subsection{Dataset}\label{sec:dataset}
The training dataset is a subset of simulations used to build Carbon's MetaMaterial library \cite{Carbon_MML}. 
To this end, we take a total of 116 high-fidelity simulations and split that into 108 training and 8 test samples. 
Every simulation step deforms the tetrahedral mesh at a rate of 20 mm/s for a total of 0.315s using 15 implicit time steps. The goal of the study is to capture up to 25\% compression, so the first 12 timesteps were used in the dataset. 
Due to the autoregressive architecture of the LGN, each timestep becomes a training point, so every simulation yields 12 unique training samples.
Sample choices are for primary cell types, i.e., pure Tetrahedral, Kagome, etc \cite{Carbon_MML}, with varying strut diameters and cell sizes (see \Cref{fig:lattices}). 
Note that this is a relatively small training set; therefore, a small fraction of the dataset was used for testing. In the following sections, we address how we use these simulations to train LGN-i and LGN-ii.

\paragraph{LGN-i Training Data}
As noted earlier, LGN-i learns the reduced (beam) representation of the tetrahedral mesh. 
Therefore, a mapping of the tetrahedral mesh to its reduced beam representation is created by using the skeletal representation. 
% by assigning each node of the tetrahedral mesh to the closest node in the strut mesh. 
The displacement components $\delta{\bs{u}}$ as well as ${I}_1$ and ${J}_2$ stresses of the tetrahedral mesh are mapped onto the nodes on the reduced mesh using the closest neighborhood interpolation.
% node to estimate the state of the strut mesh nodes.
Inverse distance weighting
is used for interpolation, so tetrahedral nodes that are farther away from the strut node will contribute less to the strut node features.

% \subsubsection{LGN-ii Training Data}
\paragraph{LGN-ii Training Data}
LGN-ii learns to map reduced representation back onto the tetrahedral mesh, as shown in \Cref{fig:LGNii_arch}.
Since the inverse mapping is on repetitive geometries and connectivities, we found a much smaller number of training samples is needed to achieve the desired accuracy.
% The training dataset for the TGN used a small subset of the simulations from the overall dataset. 
Hence, a subset of strut nodes and their surrounding tetrahedral nodes were taken from 10 simulations in the training data, which amounts to 10,000 samples across the 12 time steps. 

% \subsubsection{Data Augmentation}
\paragraph{Data Augmentation}
Augmentation on the dataset was done both in the preparation phase and stochastically during the training. 
For each simulation, multiple reduced representations were prepared with strut lengths of 1.25, 1.5, 1.75, and 2.0 millimeters to impose mesh invariance.
Stochastic augmentation during training was done by randomly rotating each training sample around the Z-axis, to make the model rotation invariant.
Random noise ($\sigma$ = 0.003) was also applied to all node positions and displacements.

\subsection{Prediction of Lattice Deformation}

\begin{table}[]
\centering
\caption{A summary of the lattices used for testing, their descriptors, and the prediction errors at 25\% strain. The unit cell composition is parameterized from a set of 7 possible unit cells\cite{Carbon_MML}. The cell size pertains to the size of the repeating unit cell.}
\label{tbl: test_set}
%\begin{tabular}{|llll|ll|}
\begin{tabular}{|p{1cm}| p{4cm}| p{1.5cm}| p{1cm}|| p{3.25cm}|p{3.25cm}|}

\hline
\textbf{Test ID} & \textbf{Unit Cell Composition}                            & \textbf{Lattice Thickness} & \textbf{Cell Size} & \textbf{Average Point by Point Error   (mm)} & \textbf{Max Point by Point Error (mm)}   \\ \hline
T1               & 100\% Kelvin                                              & 0.6                        & 9                  & $0.313 \pm 3.80 \times 10^{-2}$ & $2.50 \pm 8.97 \times 10^{-2}$ \\
T2               & 100\% Star                                                & 0.8                        & 7                  & $0.297 \pm 1.94 \times 10^{-2}$ & $ 1.92 \pm 3.09 \times 10^{-2}$\\
T3               & 100\% Voronoi                                             & 1.4                        & 10                 & $0.412 \pm 1.17 \times 10^{-2}$ & $ 2.71 \pm 3.20 \times 10^{-1}$  \\
T4               & 20\% Kagome, 40\% Voronoi, 20\% Kelvin, 20\% Rhombic      & 1.2                        & 12                 & $0.594 \pm 1.09 \times 10^{-2}$ & $ 2.90 \pm 1.15 \times 10^{-1}$ \\
T5               & 40\% Kagome, 20\% Voronoi, 20\% Kelvin, 20 \% Rhombic     & 1                          & 6                  & $0.302 \pm 7.84 \times 10^{-3}$ & $ 1.67 \pm 1.23 \times 10^{-1}$ \\
T6               & 100\% Kagome                                              & 1.4                        & 11                 & $0.376 \pm 9.39 \times 10^{-3}$ & $ 3.28 \pm 3.16 \times 10^{-1}$ \\
T7               & 100\% Octahedral                                          & 0.6                        & 12                 & $0.491 \pm 1.93 \times 10^{-2}$ & $4.41 \pm 4.32 \times 10^{-1}$ \\
T8               & 20\% Tetrahedral, 20\% Voronoi, 40\% Kelvin, 20\% Rhombic & 0.8                        & 9                  & $0.330 \pm 1.09 \times 10^{-2}$ & $2.57 \pm 1.51 \times 10^{-1}$ \\ \hline
\end{tabular}
\end{table}

\begin{figure*}[h!]
    \centering
    \includegraphics[width=\columnwidth]{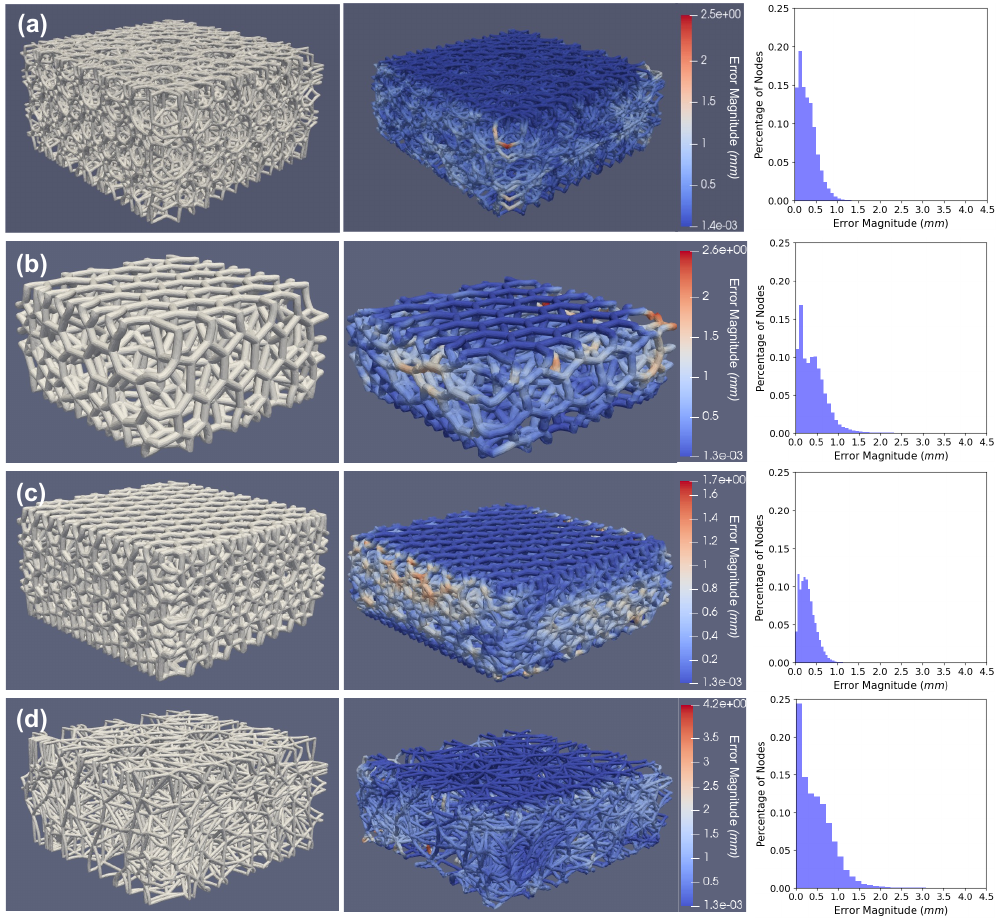}
    \caption{Visualization of the (Left) original latticed puck, (Middle) predicted deformation at 25\% compression from the LatticeGraphNet, and (Right) the distribution of displacement errors (mm) of volumetric nodes. Results are given for a) T1, b) T3, c) T5, and d) T7.}
    \label{fig:deformation}
\end{figure*}

\Cref{fig:deformation} shows multiple examples of lattice structures and the predicted deformation, along with distributions of point-by-point errors in the test set, detailed in Table \ref{tbl: test_set}.
We find that qualitatively, the results are representative of 
% what one would expect with 
the deformation of elastomeric lattices. 
LGN learns to even predict the buckling of struts. This also introduces error at less-contained places like boundaries due to the non-deterministic nature of buckling. 
The point-by-point error distributions are concentrated to low errors, with most points falling within 1 $mm$ of the ground truth IPC simulation. This average error is consistent with what is demonstrated by MGNs \cite{MGN}.
A small portion of tetrahedral nodes, mostly located at the boundaries of the lattice structure, show high errors, as highlighted by the red areas on the predicted puck deformations. { This accuracy is maintained while keeping inference times for LGN-i to ~1s and LGN-ii to ~5-7s per time step, which are a significant speed-up compared to the FEM simulations.} Note that in this error analysis, we consider different discretization lengths for the reduced representation, i.e., 1.25 mm, 1.5 mm, 1.75 mm, and 2.0 mm, to avoid any biased observations.

%\hl{maybe mention this is consistent with results from meshgraphnet? ...}

\subsection{Force Predictions}

\begin{figure*}[h!]
    \centering
    \includegraphics[width=\columnwidth]{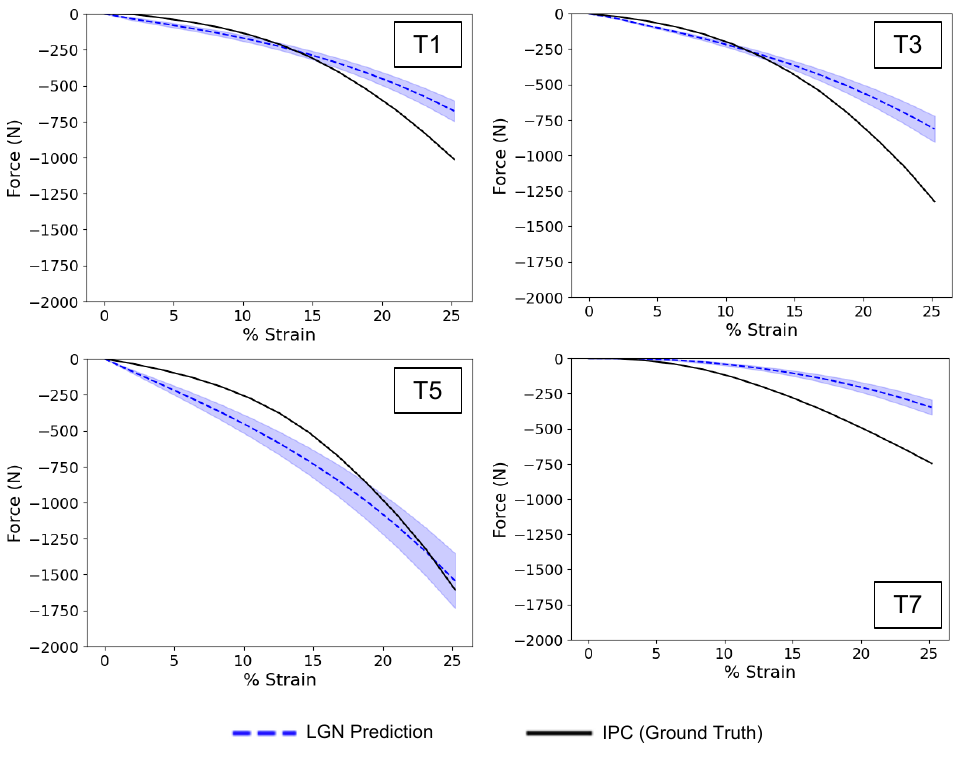}
    \caption{Feedback force of samples T1, T3, T5, T7 from LatticeGraphNet (LGN) compared with the force of the ground truth simulated sample. The shaded blue area represents the standard deviation of the predictions, caused by using multiple discretization lengths. }
    \label{fig:feedback_force}
\end{figure*}

As mentioned above, unlike FEM, LGN cannot directly predict reaction forces. To this end, we use the homogenization approach, detailed earlier, as shown in \Cref{fig:feedback_force}. While the trend of increasing magnitude of the force is captured by LGN, we see notable differences in feedback force starting at 10\% strain. The main cause of this difference is likely inaccuracies in the reduced displacement predictions propagated further by the LGN-ii. Even though the point-by-point error of most LGN predictions is less than 1 mm (\Cref{fig:deformation}), accumulation of errors could cause significant distortion on individual tetrahedrons and amount to a large amount of noise for calculating the deformation gradient and, therefore, stresses. Despite inaccuracies, we find that overall trends within the dataset are identified. For example, in \Cref{fig:feedback_force} both IPC simulations and LGN predict a high magnitude feedback force for T5, and the lowest magnitude for T7. Both models also predict that T1 and T3 are comparable in stiffness, between the extremes of T1 and T7. This demonstrates that the importance of topological features, such as strut diameter and density of structures, are captured, therefore, LGN can be used in finding potential candidate lattices for different purposes. 

% \sn{What are T1, T2, .., T7?}

\section{Conclusion}

In this work, we introduce LatticeGraphNet (LGN) for predicting the response of lattices and structures. The pipeline harnesses two Graph Neural Networks (GNNs), LGN-i for learning reduced responses of the lattices in meta-materials and LGN-ii for learning three-dimensional responses. Our pipeline is trained on a set of high-fidelity nonlinear finite-element simulations and evaluated by studying the compressive response of lattices on unseen geometries. 

To our knowledge, this is the first study that tackles the three-dimensional response of lattices with nonlinearity and buckling considerations. We find that our approach provides reasonable responses until 25\% compressive strain and can predict nonlinear behavior such as buckling. Beyond 25\%, one needs to account for self-contact, which can be added to LGN-i. 
Homogenization of the LGN outputs showed similar force correlations with IPC but contained inaccuracies. Note that the size of training simulations is relatively small compared to original MGN works; therefore, we primarily associate the observed inaccuracies with the size of the dataset. Extended training of the networks can also help improve the accuracy of the LGN network. 

{3D printing today has matured to be able to mass produce customized parts. Many of the key applications of elastomeric 3D printed products are seen in customized consumer products such as midsoles for footwear, protective pads for helmets, cushioning for bike seats, and several others. These energy-absorbing products are typically designed using expensive prototyping and physical testing cycles due to lack of simulation tools. The impact of this work is far-reaching because it offers a promising avenue to accelerate high-fidelity analysis of such products and other complex structures and drive the wider adoption of 3D printing. The differentiable nature of LGN provides immense power for tuning and optimizing lattices and structures to desired needs. These strides in digital characterization mark a pivotal advancement in the widespread adoption of machine learning technologies in {analyzing structures}. }

\section*{Acknowledgement}
We thank M.A. Nabian for helping us with setting up Nvidia Modulus. We thank Carbon Inc. for their support of this work. We additionally thank Greg Dachs and Ruiqi Chen, Carbon Inc., for their constructive feedback.

\appendix

\section{LGN and Homogenization Details}
\begin{figure*}[h!]
    \centering
    \includegraphics[width=\columnwidth]{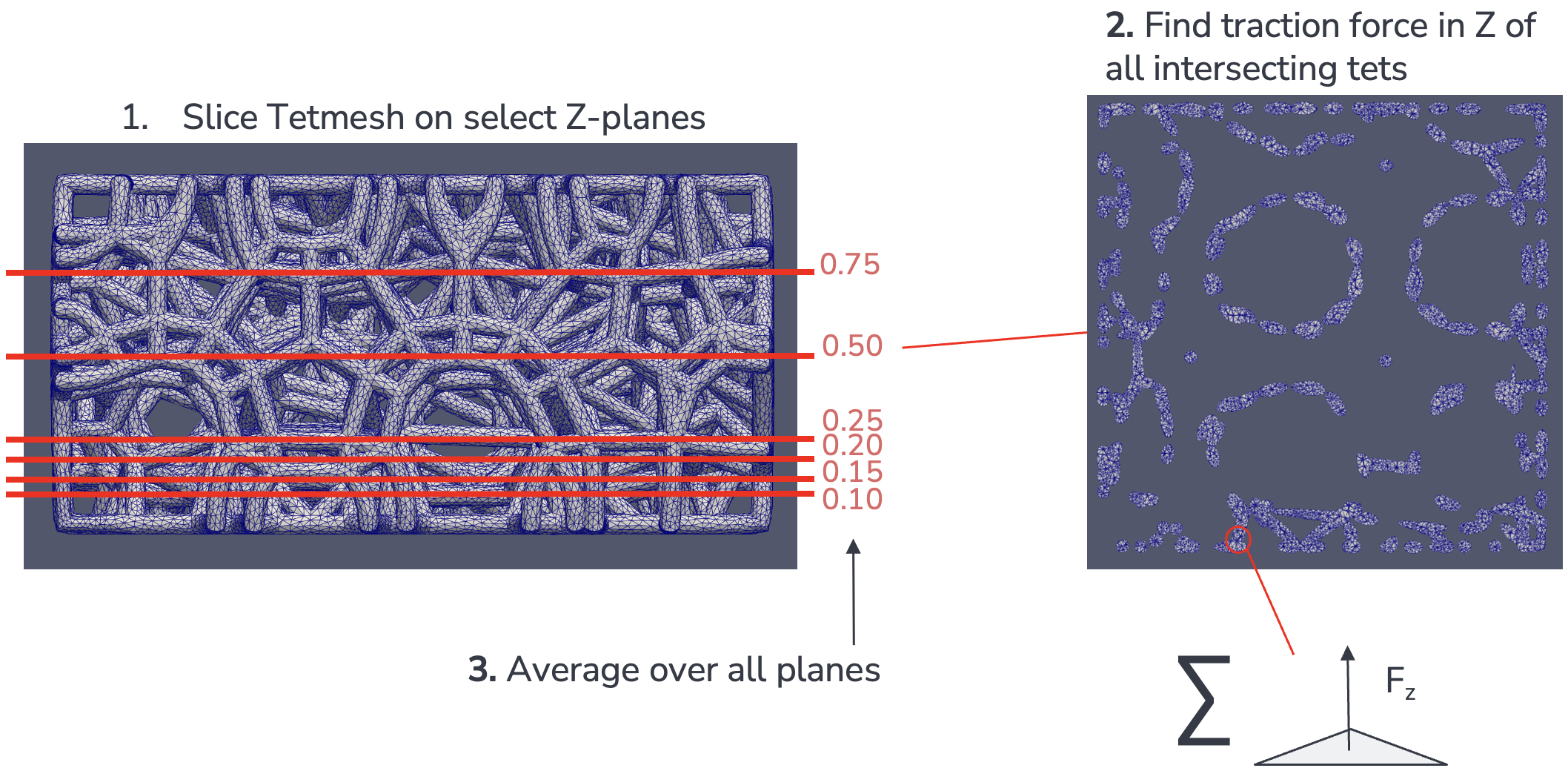}
    \caption{Visualization of the homogenization steps to determine feedback force. 1. From the initial state of the structure, we take slices at fixed heights and record all tetrahedrals that intersect with that slice. 2. We calculate the stress of every intersecting tetrahedral using \Cref{eqs:pk-stress}. The area of intersection is used to calculate the traction force $F_z$ on the area of intersection. The sum of traction forces is taken over all intersecting cells to get the total force on the plane. 3. The forces on every plane are averaged to get the feedback force of the lattice.}
    \label{fig:Homogenization}
\end{figure*}

The hyper-parameters of the LGN architecture are detailed in \Cref{tab:hyperparameters1,tab:hyperparameters2}. The homogenization procedure is depicted in \Cref{fig:Homogenization}.

\begin{table}[h!]
\centering
\begin{tabular}{|l|l|}
\hline
\textbf{Parameter Name} & \textbf{Value} \\ \hline
Initial Learning Rate           & 0.0001           \\ \hline
Decay Rate           & 0.999995          \\ \hline
Batch Size              & 1             \\ \hline
Optimizer               & Adam           \\ \hline
Activation Functions    & Hyperbolic Tangent          \\ \hline
Encoder, Decoder, Processor Hidden Layer Size          & 128              \\ \hline
Encoder, Decoder, Processor Hidden Layer Number       & 2            \\ \hline
Message Passing Steps      & 15            \\ \hline
Training Dataset Size     & 1296            \\ \hline
Training Steps     & 200,000           \\ \hline
\end{tabular}
\caption{LGN-i Hyperparameters}
\label{tab:hyperparameters1}
\end{table}

\begin{table}[h!]
\centering
\begin{tabular}{|l|l|}
\hline
\textbf{Parameter Name} & \textbf{Value} \\ \hline
Initial Learning Rate           & 0.0005           \\ \hline
Decay Rate           & 0.99999          \\ \hline
Batch Size              & 256             \\ \hline
Optimizer               & Adam           \\ \hline
Activation Functions    & Hyperbolic Tangent          \\ \hline
Encoder, Decoder, Processor Hidden Layer Size          & 64              \\ \hline
Encoder, Decoder, Processor Hidden Layer Number       & 2            \\ \hline
Message Passing Steps      & 2            \\ \hline
Training Dataset Size     & 120,000           \\ \hline
Training Epochs    & 300         \\ \hline
\end{tabular}
\caption{LGN-ii Hyperparameters}
\label{tab:hyperparameters2}
\end{table}

\newpage
\bibliography{references}

\end{document}